# Image-based Indian Sign Language Recognition: A Practical Review using Deep Neural Networks

**Mallikharjuna Rao K[1*]**
Assistant Professor
**rao.mkrao@gmail.com , mallikharjuna@iiitnr.edu.in**

https://orcid.org/my-orcid?orcid=0000-0002-6096-3963

**Harleen Kaur[2]**
**harleen20102@iiitnr.edu.in**

**Sanjam Kaur Bedi[3]**
**sanjam20102@iiitnr.edu.in**

**M A Lekhana[4]**
**mysore20101@iiitnr.edu.in**

[1*, 1, 2] Data Science and Artificial Intelligence , [3]Electronics and Communication Engineering
International Institute of Information Technology Naya Raipur, India
*corresponding author

**Abstract-** People with vocal and hearing disabilities use sign language to express themselves using visual gestures and signs. Although sign language is a solution for communication difficulties faced by deaf people, there are still problems as most of the general population cannot understand this language, creating a communication barrier, especially in places such as banks, airports, supermarkets, etc. [1]. A sign language recognition(SLR) system is a must to solve this problem. The main focus of this model is to develop a real-time word-level sign language recognition system that would translate sign language to text. Much research has been done on ASL(American sign language). Thus, we have worked on ISL(Indian sign language) to cater to the needs of the deaf and hard-of-hearing community of India[2]. In this research, we provide an Indian Sign Language-based Sign Language recognition system. For this analysis, the user must be able to take pictures of hand movements using a web camera, and the system must anticipate and display the name of the taken picture. The acquired image goes through several processing phases, some of which use computer vision techniques, including grayscale conversion, dilatation, and masking. Our model is trained using a convolutional neural network (CNN), which is then utilized to recognize the images. Our best model has a 99% accuracy rate[3].

***Keywords**:* Indian Sign language (ISL), Sign language Recognition(SLR), CNN, American Sign Language (ASL), Web Camera, Hand movements.

## 1. Introduction

According to a survey, about 15-20% of the population in the world is deaf and dumb. India, the second largest country, is home to approximately 63 million people from the DHH (Deaf and hard of hearing) community. For individuals who are deaf or mute, Sign Language is a crucial aspect of their daily existence. They rely on it daily to communicate with their peers[1]; the complexity of these signs and the knowledge of these gestures is unknown to many, essentially hampers their communication with the real world. The progress made in deep learning and computer vision can resolve the issue. Using Indian sign language, [1] aims to bridge the communication divide between individuals who can hear and those who are deaf. Enlarging the scope of this initiative to include characters, words, and everyday phrases can not only enhance the interaction between deaf and mute individuals and the external environment. However, they can also promote the development of self-governing systems capable of understanding and assisting them[2].

Three sign language recognition approaches can be used- Character- level, i.e., translating every gesture into the character it denotes; word level, i.e., translating the gestures into the respective word; and Sentence-level, i.e., translating the gestures into an entire sentence. In our model, we have implemented character-level sign language recognition and analyzed it with different deep learning and machine learning models to find the best model suitable for such problems. Also, we did some work to solve for real-time word-level recognition[3].

Sign language is not universal. People from different countries use different sign languages. However, the existing research emphasized American Sign Language (ASL), which needs to be more accurate for India [4]. Thus, this model focuses on Indian Sign language so that the deaf and mute people in India can easily use it. The significant differences between ASL and ISL are 1. ASL involves one-handed symbols, while ISL involves two-handed symbols. 2. Classification of ISL is difficult due to finger angle variations, whereas for ASL, it is simple. 3. In ISL, most symbols are almost similar, whereas, in ASL, symbols are easily distinguishable[4]. Because of the intricate hand gestures utilized in Indian sign language, relatively little research has been conducted.

This paper aims to demonstrate how to recognize the Indian Sign Language alphabet using the corresponding gesture. The identification of gestures and sign language has been extensively studied in American Sign Language, but it has received limited attention in the context of Indian Sign Language. To tackle this issue, we suggest recognizing movements from images captured by a camera rather than relying on technologies such as gloves or Kinect. Subsequently, we will use computer vision and machine learning approaches to extract and classify specific features [1].

Therefore, the primary objectives of this research are

1. Generating an ample amount of dataset for the Indian sign language
2. Designing the best model for CNN to train the preprocessed images and achieve the maximum possible accuracy.

This research is organized as follows: section 2 summarizes the performed literature study, and section 3 discusses deep learning architecture models. Section 4 of the paper discusses the results achieved using our methodology, and section 5 discusses the conclusions we have arrived at from this research project.

## 2. Literature Review

A systematic and detailed literature review was conducted to proceed with this research project. Gautham Jayadeep et al. [1] proposed a solution Mudra, a sign language translation tool for vocally disabled people in the banking sectors. It focused on collecting data on sign languages, preprocessing the data, feature extraction and training using CNN, and then the classification of signs to text. The major limitation is that the dataset used is simple, whereas any regular sign language has complex hand signs. The background and hand contact with the body has yet to be considered a factor in this proposed solution[1].

Ismail Hakki Yemenoglu et al. [2] did research on a "CNN-based sign language recognition system." The research was done on the American Sign language dataset, and it focused on Character-level sign language recognition. This research used a dataset comprising ten samples of each sign letter, with each image measuring 224 by 224 pixels. The proposed approach employed a CNN called GoogleNet, which can categorize thousands of objects. To adapt GoogleNet to recognize sign language, the researchers utilized a transfer learning technique and conducted their experiments using MATLAB. The resulting accuracy of the sign language recognition system was 91.02% [2].

In their study on "Real-time recognition of Indian sign language," Dr. Gomathi V et al. [3] utilized the Fuzzy C-means Clustering Machine Learning Algorithm to train and predict hand gestures. Although FCM was effective, it necessitated more computation time than other methods. Therefore, the researchers proposed an extension of the system that integrates Convolutional Neural Networks (CNN) and Recurrent Neural Networks (RNN) to capture spatial and temporal features[3].

Mohammed Safeel et al. [5] conducted research to evaluate the many SLR techniques that have been used recently. They have studied various images based on or without gloves for detection and studied their advantages and difficulties during the process. Training models like Hidden Markov model-based approaches and Deep learning methods like CNN, KNN, ANN, and SVM have also been discussed. Finally, the challenges and a comparative study of all the proposed models were discussed[5].

In their research on "Indian Sign Language Recognition System using SURF with SVM and CNN," Shagun Katoch

et al. [6] introduced a variety of visual words model to identify characters in Indian sign language ranging from 0 to 9 and A to Z. SVM and CNN was used for classification. An analysis is done between the two approaches. This proved to be a better model than where HoG and SVM were used for feature extraction, as it performed better in accuracy[6].

Thus, many concerns were observed during the literature review. Most of the studies focused on the ASL dataset. Some studies used Character-level recognition, while some focused-on Word-level. No study was done on the sentence level as it is complex and inaccurate. Regarding training the model, different studies focused on algorithms such as fuzzy means, CNN, SVM, etc. Therefore, considering the literature study, this proposed solution focuses on implementing CNN-based Character-level Indian sign language recognition and a comparison with the existing solutions.

## 3. Proposed Solution

Our study utilized the publicly available ISL-Character dataset comprising Indian Sign Language (ISL) characters. We used various preprocessing techniques to improve the quality of images, removing undesired objects to keep the focus on the hand and then training our dataset on various models to do a comparative analysis for ISL. Fig 1. shows the proposed model diagram.

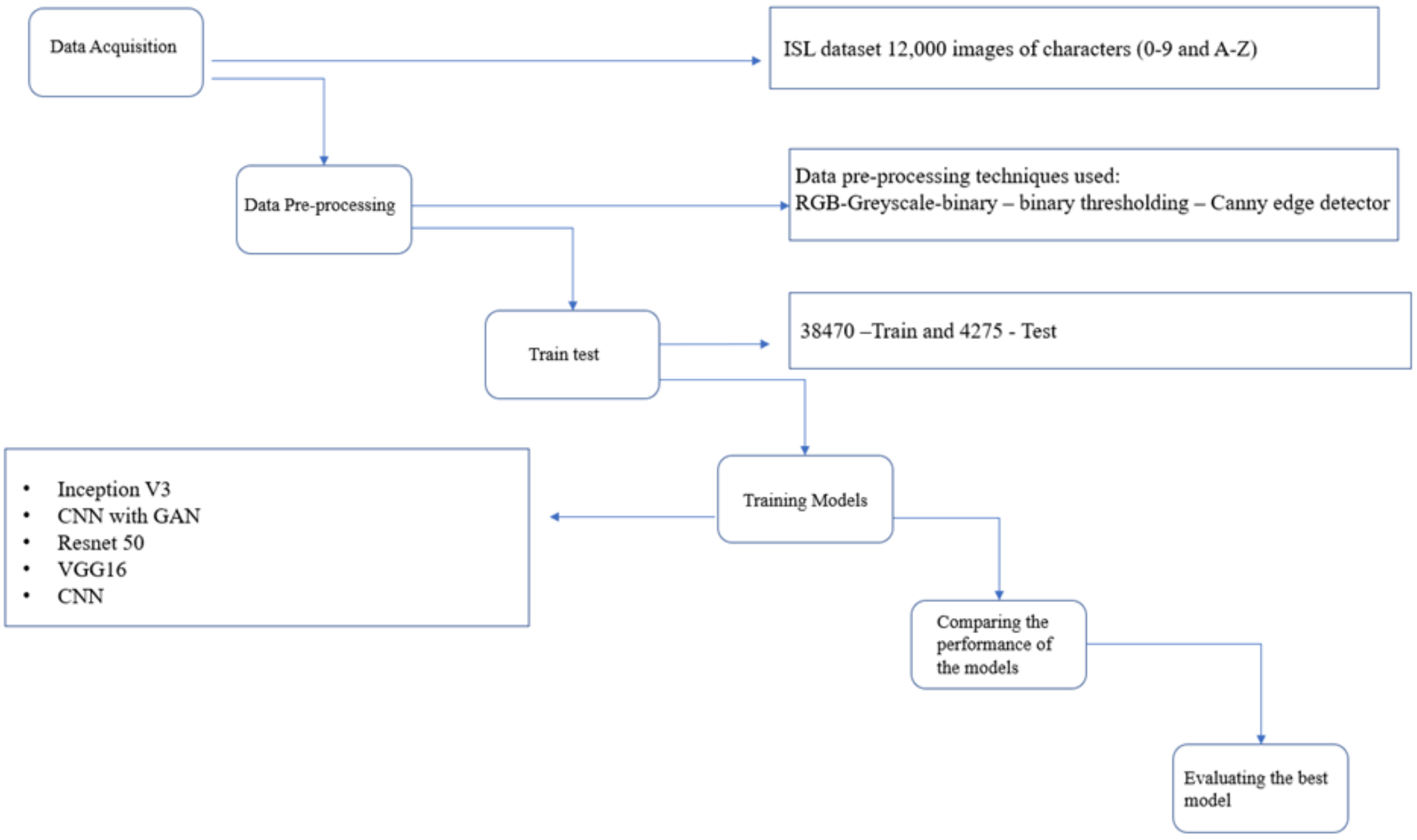

Fig 1. Proposed Model

### 3.1. Data Acquisition

The ISL- Character dataset is publicly available for Indian sign language. It consists of 36 characters from 0 to 9 and A to Z. Each character consists of 1200 images denoting that particular character, which have been taken in various alignments for both left and right hands. Samples of the ISL dataset have been shown below in Fig 2.

### 3.2 Image Preprocessing

A crucial stage in computer vision preprocessing is image resizing. All of the photos have the exact dimensions to maintain scale homogeneity. Because of this reason, all the images present in the dataset are resized to 226x226. To simplify the training and recognition process, all the images are converted to grayscale so that only the critical features can be focused on [9].

Most of the images were either very bright, underexposed, or blurry. The noise from these photos then hampers the model. As a result, the model's output must reflect how well it performs, accurately necessitating dataset cleansing. We used the Gaussian blurring technique to remove the noise present in the image dataset and for smoothening. This is followed by the median filter, which removes the salt and pepper noise in the dataset. Fig 3. shows the conversion from RGB to grayscale[9]. Contrast adjustment can be seen in Fig 4.

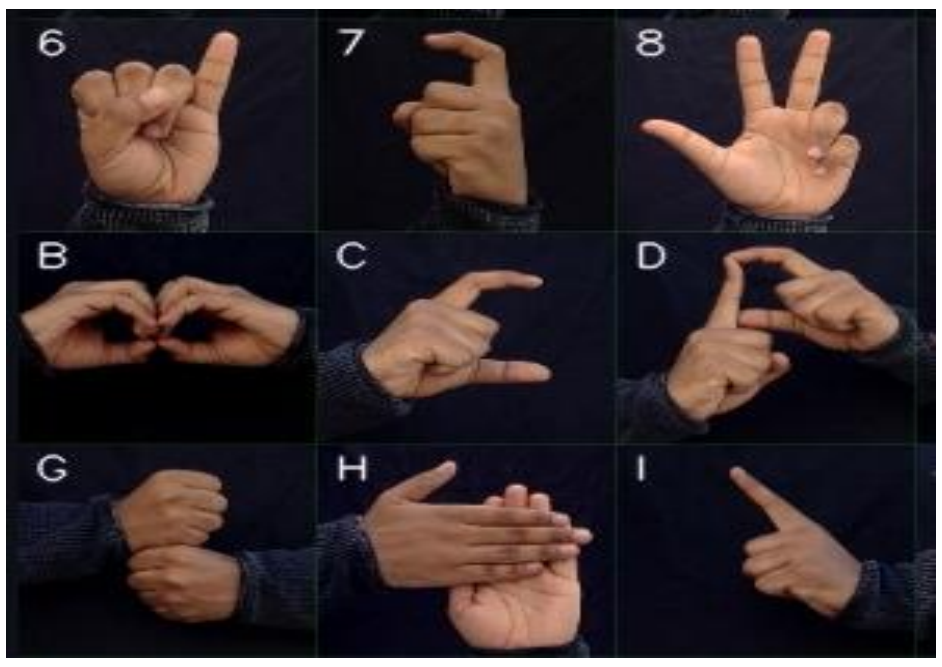

Fig 2. Samples of ISL Dataset

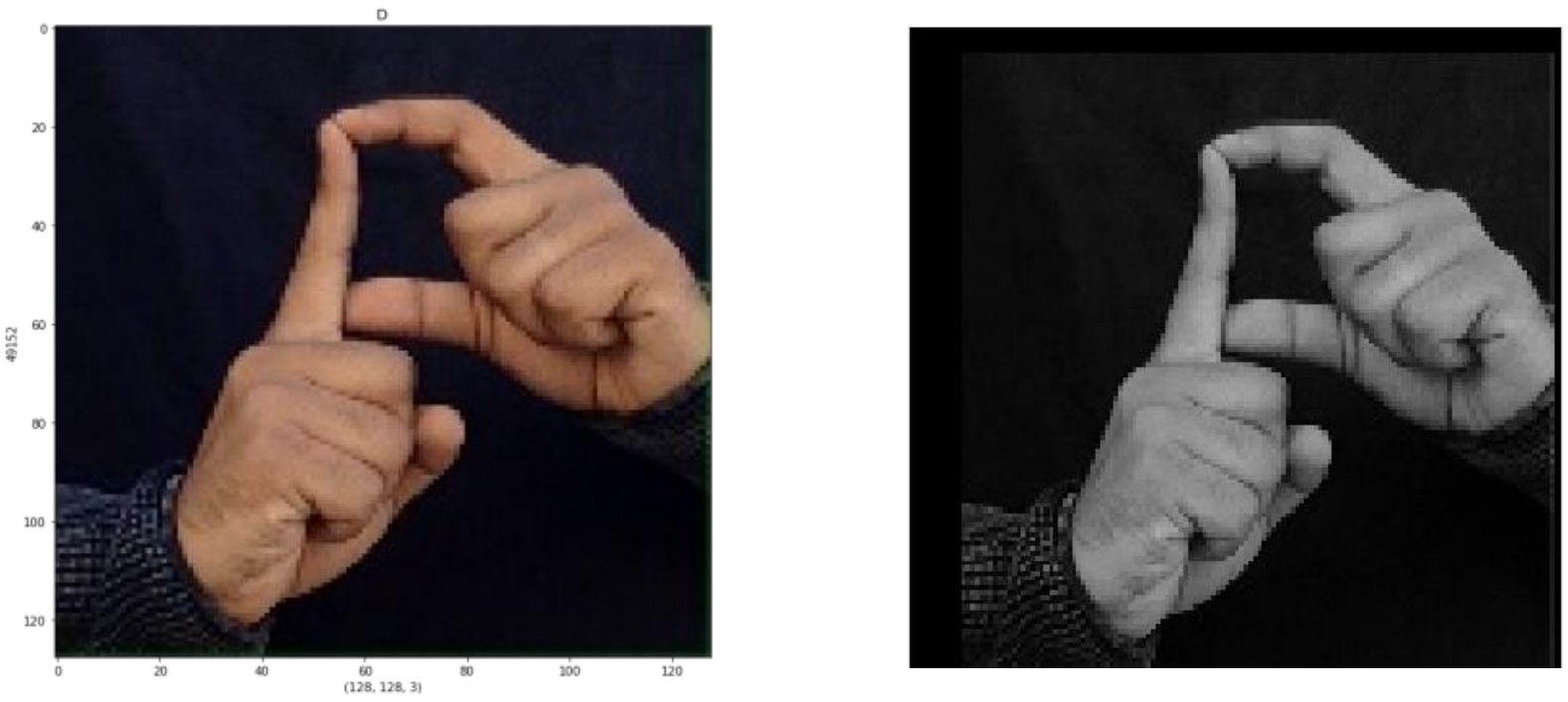

Fig 3. Converting RGB to grayscale

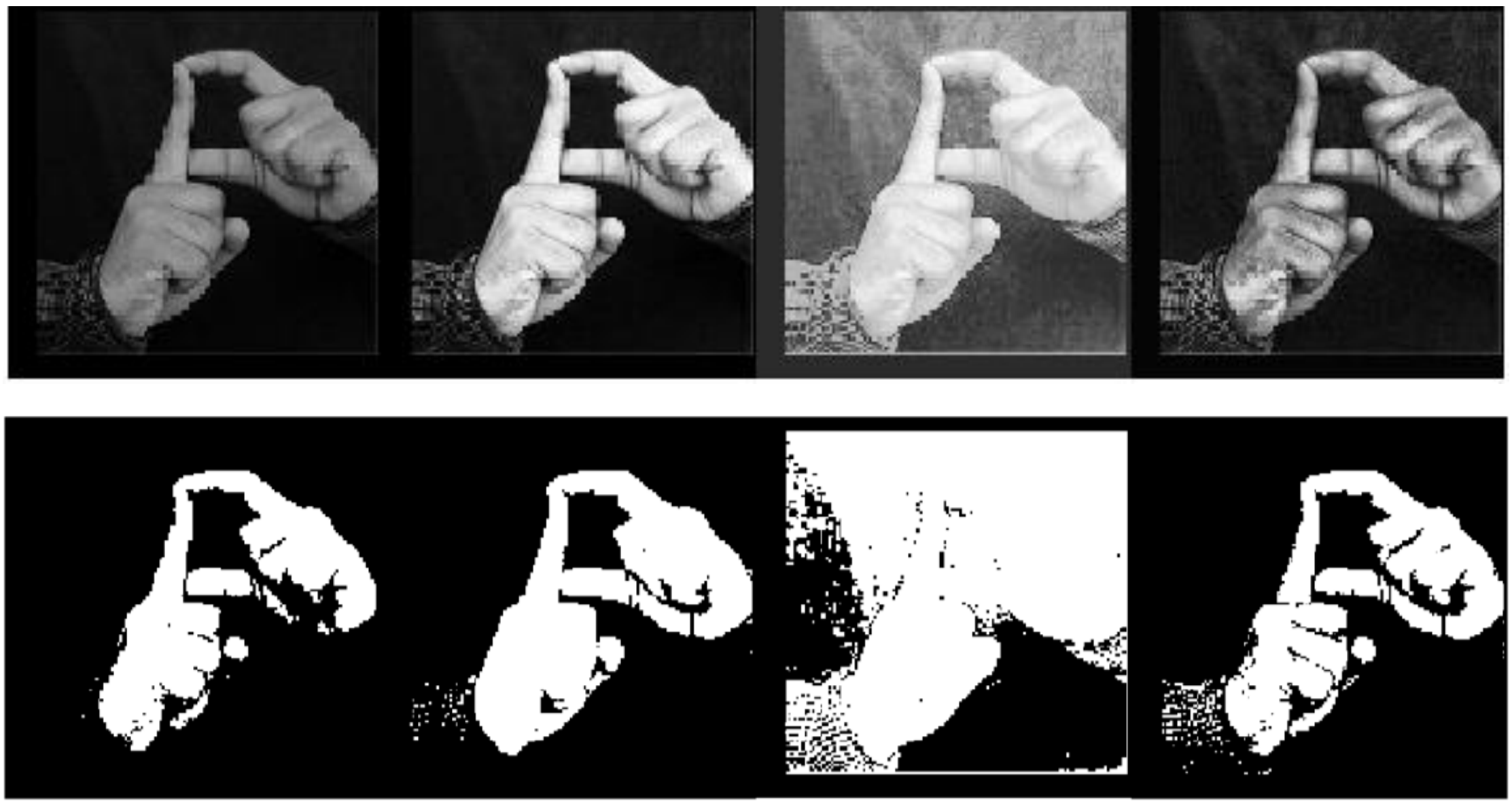

Fig 4. Contrast Adjustment

### 3.2.1 Image Segmentation

An image segmentation technique divides a digital image into several groupings, or "image segments," each containing a variety of the image's elements. Focusing on a specific item inside an image is done by image segmentation, which is necessary and helpful for our processing. Image segmentation simplifies image processing and analysis while also bringing down the complexity of the original image. In segmenting a picture, each pixel is given a name, and a section of the image is allocated to a category with a specific label. Thus, a specific area of the picture with the same label is discovered instead of the entire image[9].

A mask for hand segmentation is produced by extracting the foreground area with the highest degree of connectivity and assuming it to be the hand.

**3.2.1.1 Thresholding**

Thresholding is a type of picture segmentation where the pixels are altered to make the image simpler to understand. Thresholding is a technique that converts a color or grayscale image into a binary image containing only black and white pixels. It is most frequently used to choose specific regions of an image while disregarding other regions. Suppose a pixel's intensity in the input image exceeds the threshold. In that case, the corresponding output pixel is labeled as white (foreground), and if it is equal to or less than the threshold, it is labeled as black (background)[7].

Two types of thresholding are performed:

1. Binary thresholding:

   In our model thresholded limit is set to 90 pixels for binary thresholding purposes. All the pixels having values above 90 are converted to 255, and those below 90 are converted to 0 pixels[7].

2. Otsu Thresholding:

Fig 5. shows the binary conversion after thresholding.

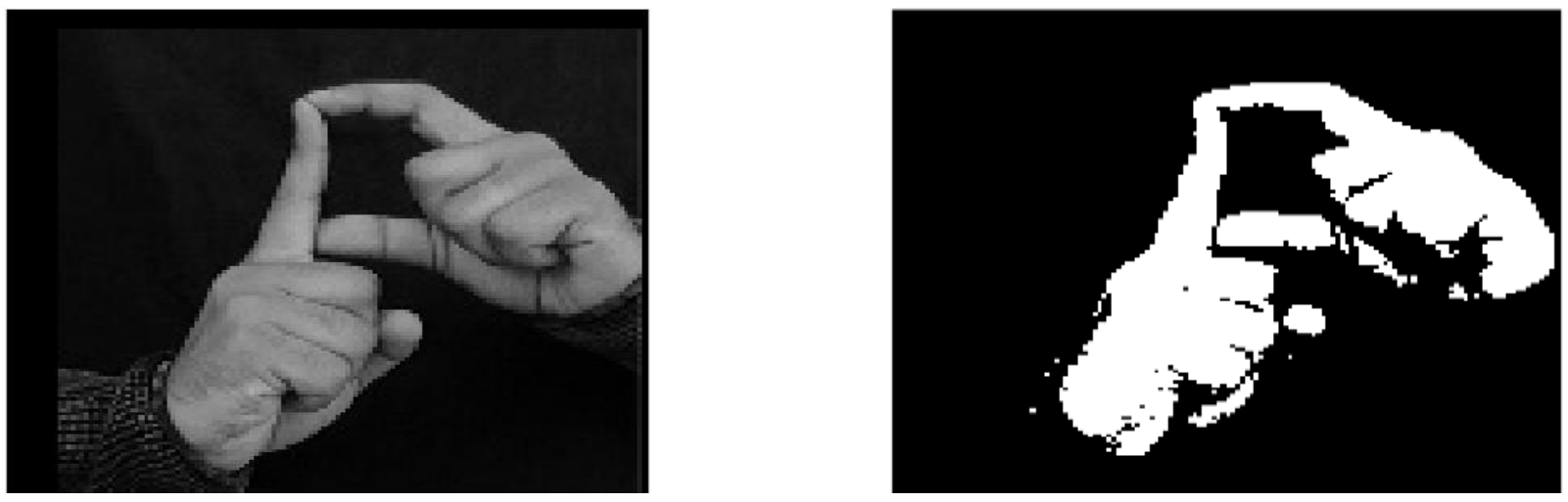

Fig 5. Binary conversion after thresholding

**3.2.1.2 Canny Edge Detector**

After that, the canny function is used to determine the intensity and direction of the pictures' edges using the gradient of each pixel. This causes an intensity change with the original picture, making the edge visible. The Canny edge detector, an edge detection operator, recognizes various image edges using a multi-stage methodology. The variation in color intensity is used to carry out edge detection. The lower and upper threshold values are set to 10 and 100[7]. Canny edge detection is shown in Fig 6.

**3.3 Feature Extraction and Classification**

The Convolutional Neural Network (CNN) is utilized for feature extraction and classification. The feature extraction phase comprises a convolution layer, a pooling layer, and an activation function, while the classification stage involves the fully connected layer. Below, we will discuss CNN and its layers.

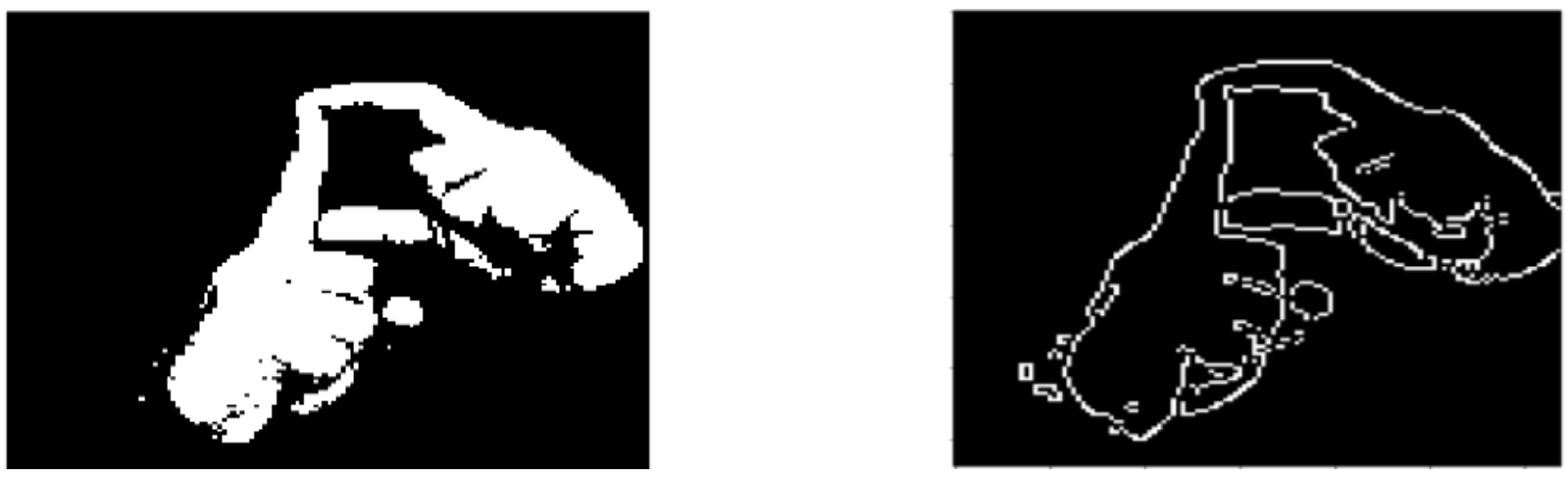

Fig 6. Canny edge detection

### 3.3.1 Convolutional Neural Network

Deep learning has become a potent tool over the last few decades due to its capacity for handling massive amounts of data. Hidden layer technology is much more popular than conventional methods, particularly for pattern recognition. Convolutional Neural Networks are among the most widely used deep neural networks. Deep learning applications often use Convolutional Neural Networks (CNN/ConvNet) to process visual data. These networks are specifically designed for this purpose. Typically, matrix multiplications come to mind when we think of a neural network, but that is not true with ConvNet. It makes use of a unique method called convolution. In mathematics, convolution is an operation that combines two functions to produce a third function that illustrates how one function modifies the shape of the other[1].

A Convolutional Neural Network typically has an input and output layer with multiple hidden layers. In a feedforward network, intermediate layers are considered hidden as the activation function, and the final convolution masks them. The convolutional neural network's hidden layers usually comprise convolutional layers, which commonly include a layer that performs a dot product between the input matrix of the layer and the convolution kernel. ReLU is often used as the activation mechanism for this product, which is frequently the Frobenius inner product. The convolution procedure develops a feature map as the convolution kernel moves across the input matrix for the layer, adding to the input of the following layer. Following this are further layers like normalizing, pooling, and fully connected layers. The CNN architecture is shown in Fig 7[1].

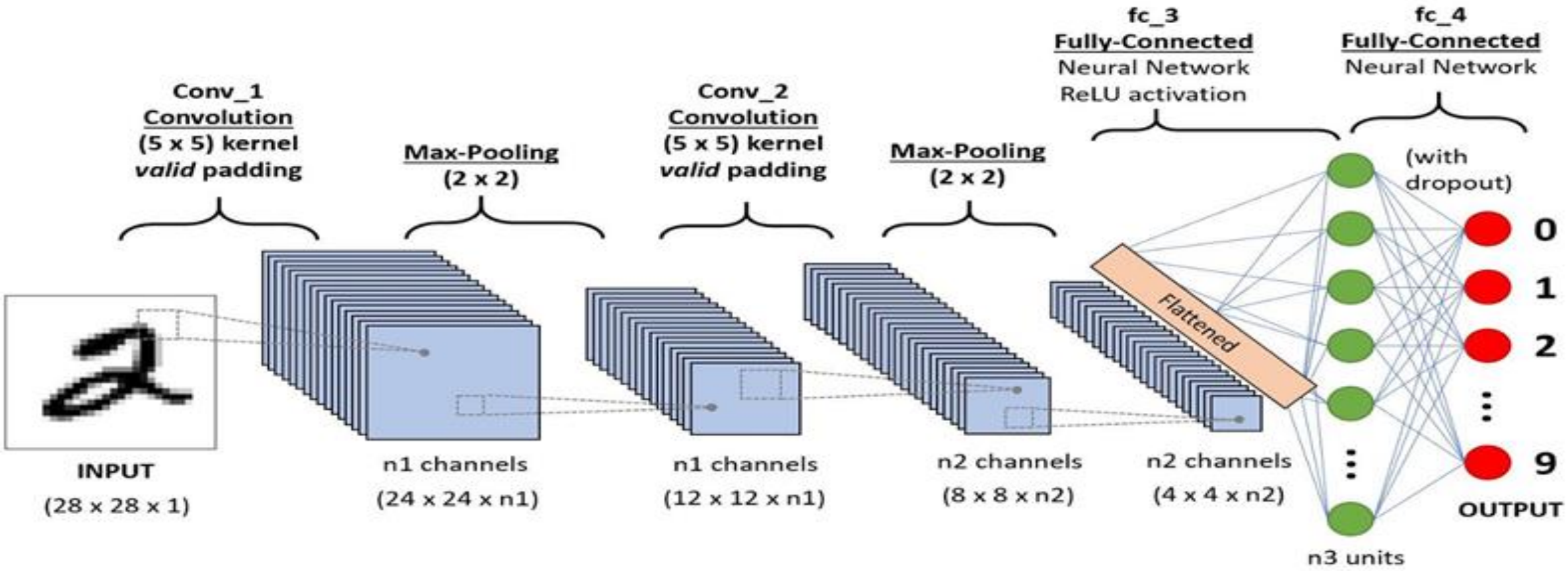

Fig 7. CNN layers

#### 3.3.1.1 Convolutional layer

In a convolutional neural network (CNN), the input is a tensor with the dimensions (number of inputs) (input height) (input width) (input channels). The input image is transformed into a feature map, also known as an activation map, with the dimensions (number of inputs) (feature map height) (feature map width) (feature map channels). The convolutional layers integrate the input and pass the resulting information to the next layer. This shows how a neuron comparable to the visual brain might react to a particular stimulus. A convolutional neural network is utilized to process data for each specific receptive field of a convolutional neuron.

On the other hand, fully connected feedforward neural networks can learn features and classify data. More significant inputs like high-resolution photos typically make this architecture unworkable. Due to the massive input size of images, where each pixel is a significant input characteristic, would require a large number of neurons, even in a shallow architecture[1].

#### 3.3.1.2 Pooling layers

Convolutional networks may include local and global pooling layers and regular convolutional layers. Pooling layers reduce the dimensionality of data by combining the outputs of neuron clusters at one layer into a single neuron at the following layer. Local pooling merges small clusters; 2x2 tile sizes are typically used. Global pooling has an impact on every neuron in the feature map. The maximum and average pooling methods are the two most used. Max pooling uses the maximum value of each local cluster of neurons in the feature map instead of average pooling, which uses the average value of each[1].

#### 3.3.1.3 Fully connected layers

Through fully connected layers, every neuron in one layer can communicate with every neuron in every other layer. The network has the same structure as a traditional multilayer perceptron neural network (MLP). The flattened matrix runs through a fully linked layer to categorize the images[1].

We have deployed four CNN architectures to train and classify our dataset, including the traditional CNN model and pre-trained models like Inception V3, ResNet-50, VGG-16, and GANs employing CNN[1].

### 3.3.2 Traditional CNN model

Our overall architecture resembles the CNN, with many dense and convolutional layers. The first layer of the design consists of a pair of two convolutional layers with a total of 32 filters and a 3x3 window size. A dropout layer and a max-pool layer are added after this. In conclusion, there is an output layer of the softmax activation function and an ultimately linked hidden layer with 512 neurons of the ReLU activation function. Two more convolutional layers with 64 filters and a maximum pooling layer are also present. A 100,100-pixel image can be fed into the first convolution layer, and the final output layer is composed of 36 neurons, one for each category of ISL indicators.

The CNN model uses Relu and Softmax as activation functions, categorical cross entropy loss function, and accuracy as metrics for the research purpose. Adam optimizer with 0.01 as the learning rate is used. The model training and testing accuracy concerning epochs is shown below in Fig 8.

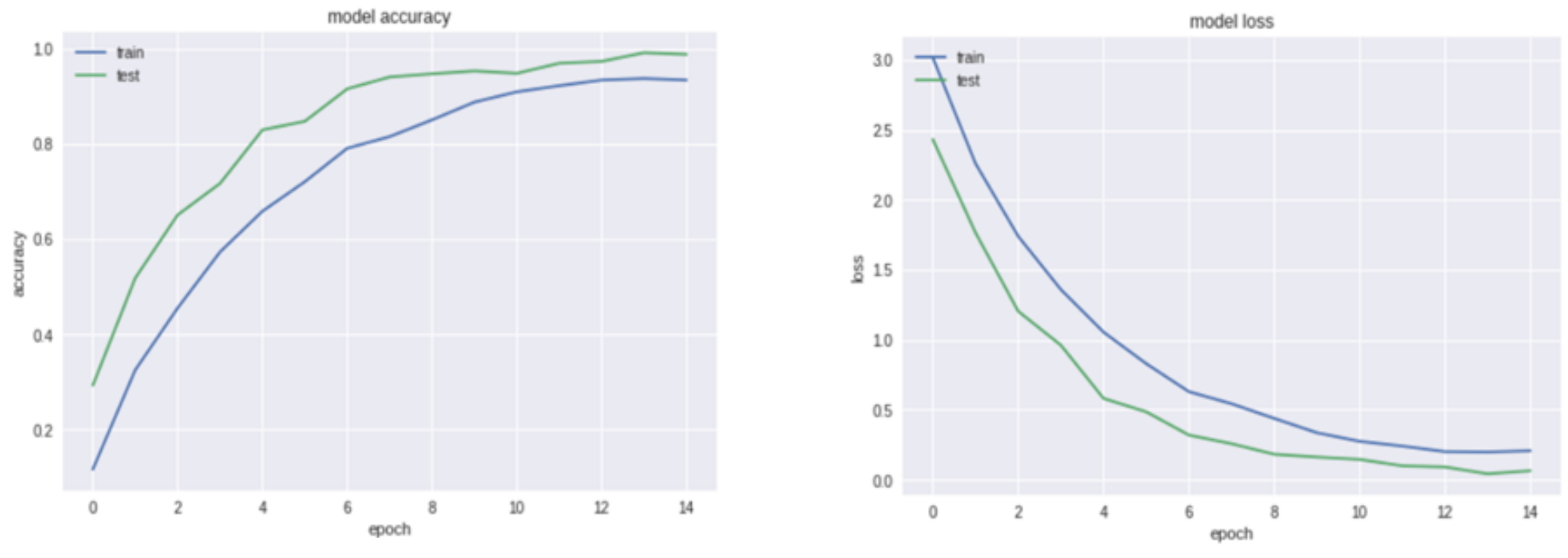

Fig 8. Training and Testing accuracy and loss

### 3.3.3 AlexNet

Convolutional neural network architecture is known as AlexNet. On September 30, 2012, AlexNet participated in the ImageNet Large Scale Visual Recognition Challenge. The network outperformed the runner-up by more than 10.8 percentage points with a top-5 error of 15.3%. The main finding of the original research was that the depth of the model, which was computationally expensive but made possible by using graphics processing units (GPUs) during training, was crucial for its high performance. Eight layers make up the architecture: three fully linked layers and five convolutional layers. However, this is not what makes AlexNet unique; instead, the following are some of the features that represent novel convolutional neural network techniques:

Nonlinearity in ReLU: AlexNet uses Rectified Linear Units instead of the tanh function, which was the industry standard then (ReLU). A CNN using ReLU was able to achieve a 25% error on the CIFAR-10 dataset six times faster than a CNN using tanh, giving ReLU the advantage over tanh in terms of training speed.

Several GPUs.: Graphics processing units (GPUs) at the time still had 3 GB of RAM (nowadays, those kinds of memory would be rookie numbers). This was not good because the training set contained 1.2 million images. By dividing the model's neurons between two GPUs, with half of them located on each GPU, AlexNet offers multi-GPU training. This shortens the training time while also enabling the training of a larger model.

Multiple Pooling: The outputs of nearby neuronal groups are often "pooled" by CNNs without any overlap. Nevertheless, when the scientists added overlap, they observed a 0.5% error drop and discovered that overlapping pooling models are typically more challenging to overfit. Fig 9. shows the architecture for AlexNet

.

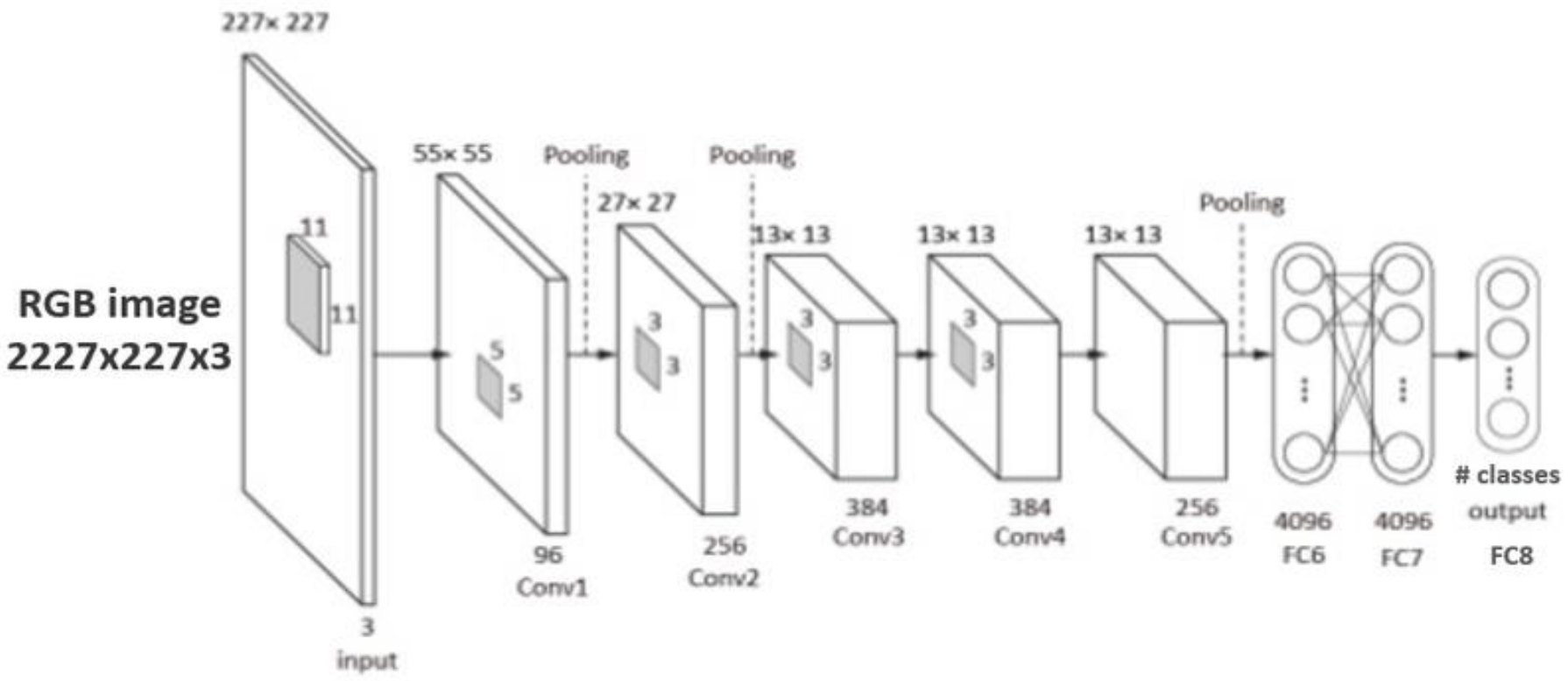

Fig 9. AlexNet Architecture

The trained Alexnet model is utilized for training with accuracy and category cross entropy loss function as metrics. It uses the Adam optimizer with a learning rate of 0.001. Fig 10 below displays the training and testing accuracy of the model about epochs.

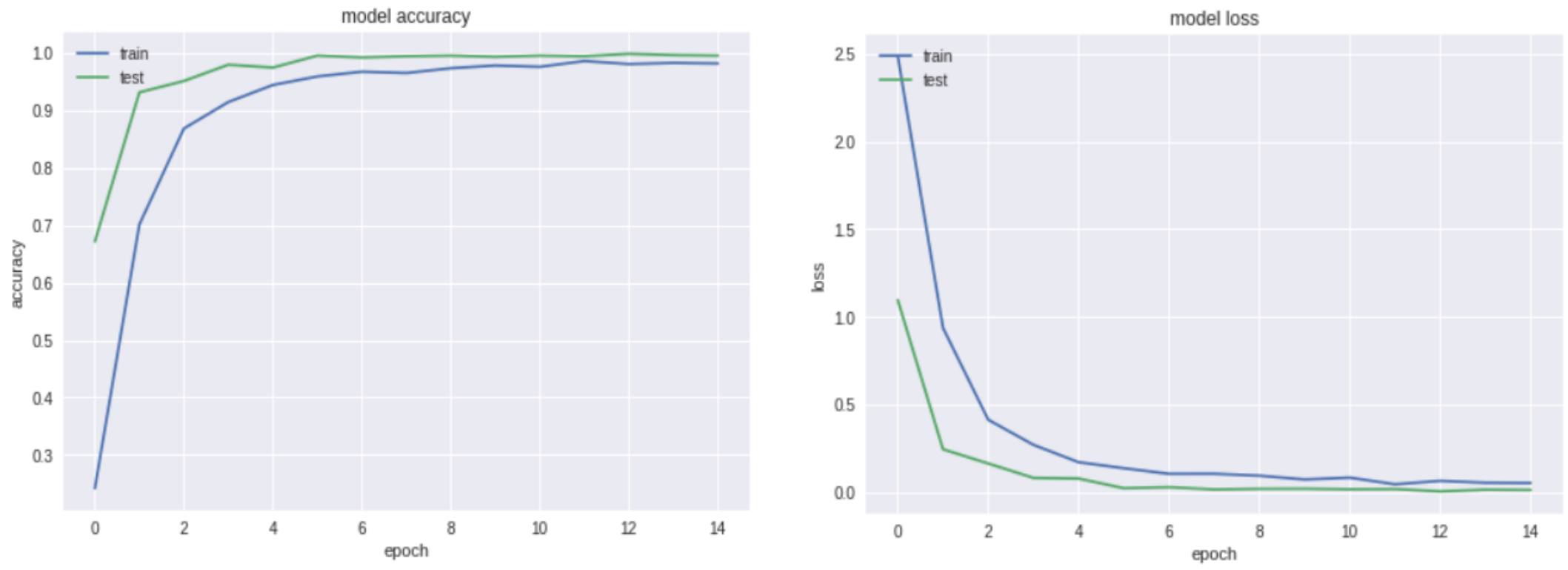

Fig 10. Training and testing Accuracy and Loss for AlexNet

### 3.3.4 InceptionV3

Convolutional neural network-based Inception V3 is a pre-trained deep learning model for image classification. The Inception V3, a core model first presented as Google Net in 2014, is an improved version of the Inception V1. The ImageNet dataset has an accuracy of over 78.1%. The model's symmetric and asymmetric building blocks include convolutions, average pooling, max pooling, concatenations, dropouts, and completely linked layers. The model heavily utilizes batch normalization, which is also used for the activation inputs. Softmax is employed to compute loss[2].

An Inception v3 network's architecture is developed gradually and methodically, as follows:

1. Factorized convolutions: This method reduces a network's computational efficiency by reducing the number of parameters. Also, it keeps an eye on the network's efficiency.

2. Smaller convolutions: Training is unquestionably sped up by swapping out larger convolutions for smaller ones. Say a 5x 5 filter has 25 parameters; replacing it with two 3x 3 filters results in only 18 (3x 3 + 3x 3) parameters[2].

3. Asymmetric convolutions: A 1x 3 convolution can be followed by a 3x 1 convolution instead of a 3x 3 convolution. If a 3x3 convolution were replaced with a 2x2 convolution, the parameters would be slightly higher than the suggested asymmetric convolution.

4. A tiny CNN is positioned between layers during training, and any losses it experiences are added to the losses experienced by the leading network. GoogLeNet uses auxiliary classifiers for a deeper network instead of Inception v3, which uses them as a regularizer.

5. Grid size reduction: Pooling techniques are typically used to reduce grid size. However, a more effective method is suggested to address the computational cost bottlenecks[2]. Fig 11. shows the Inception V3 architecture.

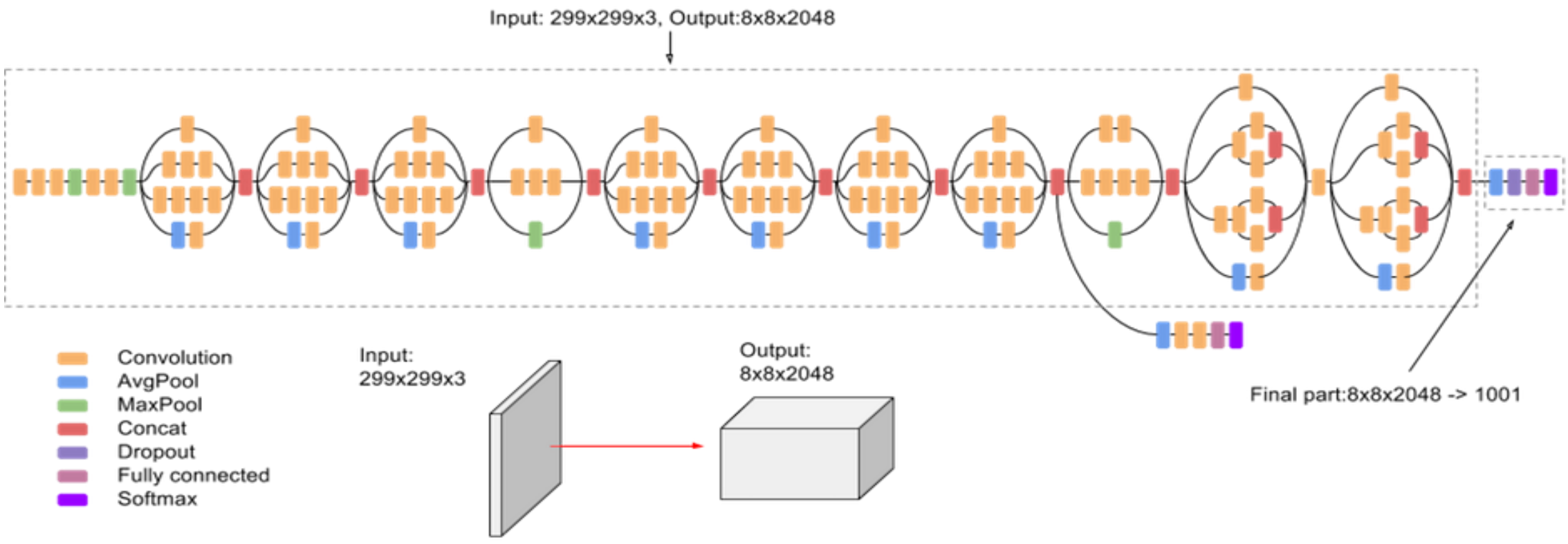

Fig 11. Inception V3 Architecture

The trained InceptionV3 model is utilized for training with accuracy and categorical cross entropy loss function as metrics. 20 iterations of the Adam optimizer with a learning rate of 0.001 are employed. Fig 12 below displays the training and testing accuracy of the model concerning epochs.

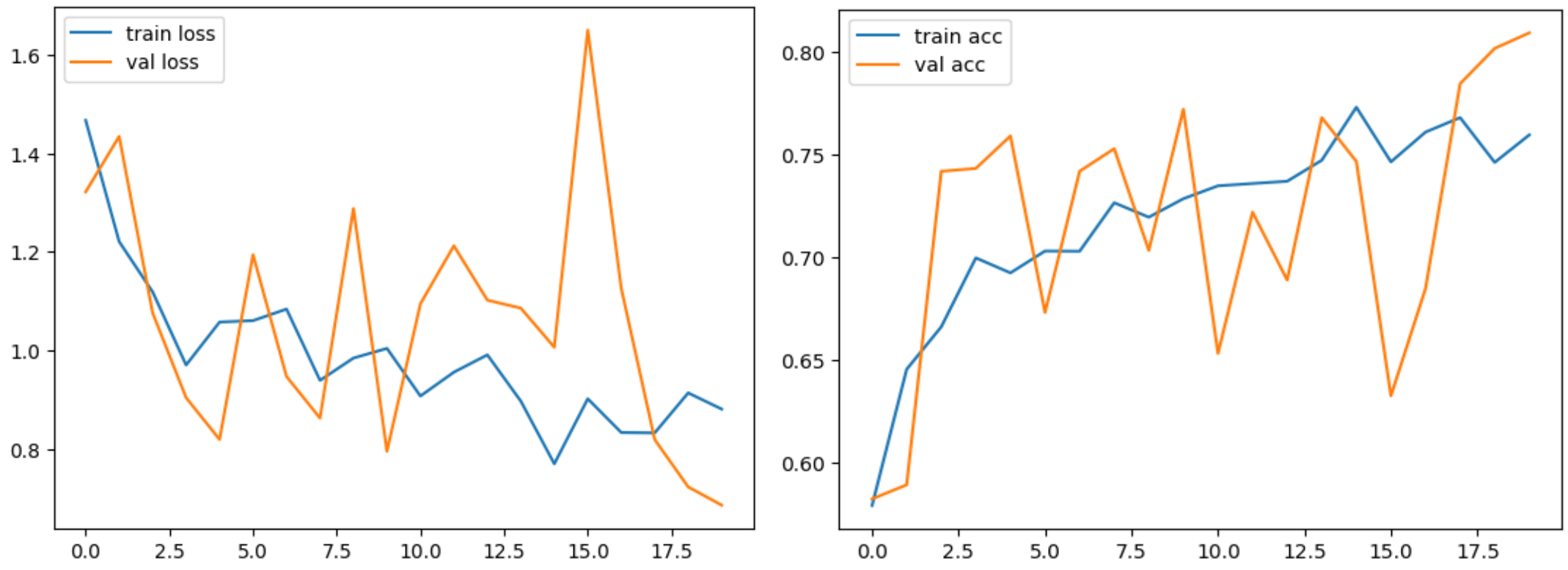

Fig 12. Training and Testing Accuracy and Loss for Inception V3

### 3.3.5 CNN with GAN (Generative Adversarial Networks)

A potent class of neural networks called GANs is employed in unsupervised learning. GANs are capable of producing anything you give them since they Learn-Generate-Improve. You must have a basic understanding of convolutional neural networks to understand GANs. CNNs are taught to categorize photos according to their labels. When a picture is input to a CNN, the CNN analyses the image pixel by pixel passes the image via nodes in the CNN's hidden layers and outputs a description of the image or what it perceives in the image[10].

The Generator and the Discriminator are the two components of GANs. Discriminator: This function of GANs is comparable to that of CNNs. The output layer of GANs can only have two outputs, in contrast to CNNs, which can have outputs proportional to the number of labels it is trained on. Discriminator is a Convolutional Neural Network composed of several hidden layers and one output layer. Due to the discriminator's carefully selected activation function for this assignment, the output can only be either 1 or 0. If the output is 1, the provided data is deemed authentic; if it is 0, it is deemed fake data[10].

The Generator is an Inverse Convolutional Neural Net; it performs the exact opposite of what a CNN does because, in a CNN, an actual image is provided as input, and a label for classification is anticipated as an output. In contrast, in Generator, random noise (expressly, a vector with some values) is provided as input, and an actual image is anticipated as output. Simply put, it uses its own creativity to generate data from a piece of data[10].

The Generator seeks to fool the discriminator, an expert who can tell real information from phony, into believing that the generated data is real. The Generator develops and grows with each failed attempt, providing more realistic data. It can also be described as a conflict between Generator and Discriminator[10].

Fig 13. shows the Generator and Descriptors in GANs.

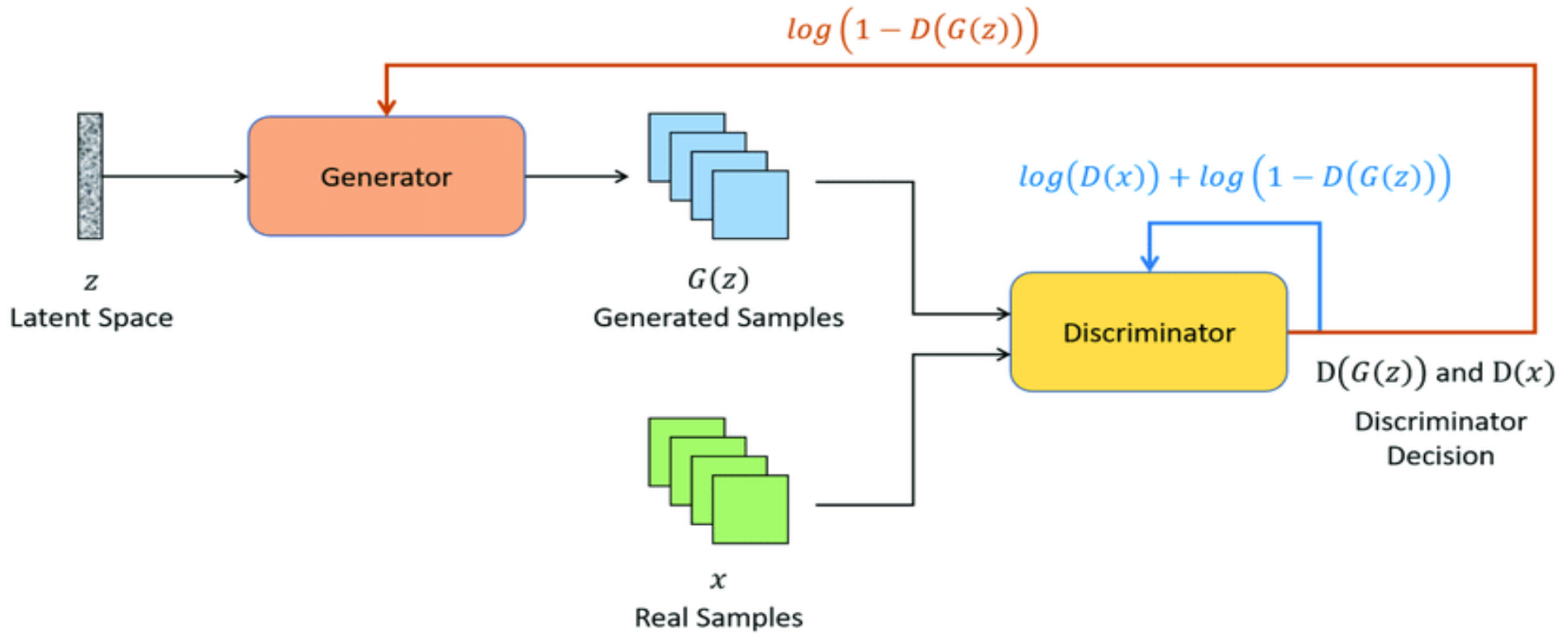

Fig 13. GANs (Generator and Discriminator)

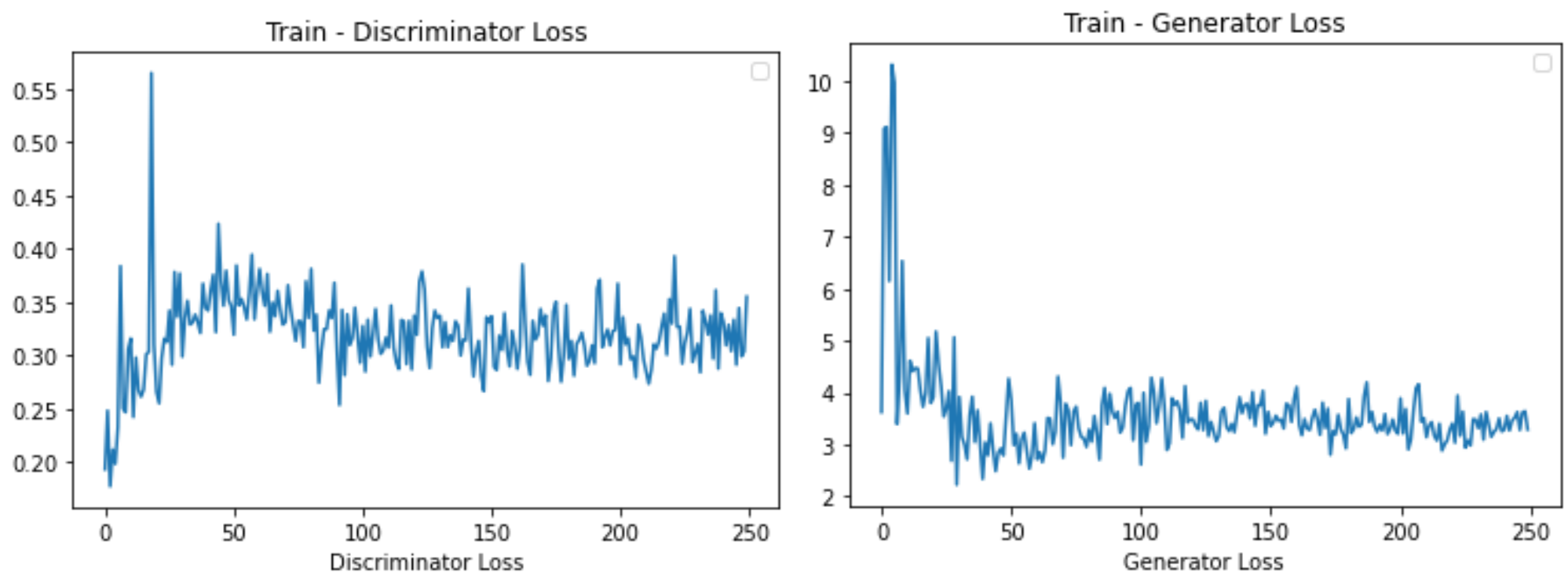

Fig 14. Discriminator and generator Loss

### 3.3.6 Resnet-50

The convolutional neural network ResNet-50 has 50 layers in total. The ImageNet database contains a pre-trained version of the network trained on more than a million photos. This design developed the Residual Blocks concept to deal with the vanishing/exploding gradient issue. In this network, we use an approach called skip connections. By passing some intermediate levels, the skip connection connects layer activations to later layers. As a result, a block is left over. The remaining blocks are piled to form resnets. This network's approach lets the network fit the residual mapping rather than having layers learn the underlying mapping. Thus, let the network fit instead of using, say, the initial mapping of H(x),

Giving rise to H(x):= F(x) + x, F(x):= H(x) - x. The advantage of this type of skip link is that regularisation will skip any layer degrading the architecture's performance. Hence, training a deep neural network is feasible without suffering from vanishing or expanding gradient problems[6].

ResNets are more understandable than traditional neural network topologies. The images of a 34-layer plain neural network, a 34-layer residual neural network, and a VGG network are shown below. The layers in the plain network have the same number of filters for the same output feature map. If the output feature sizes are reduced by half and the number of filters doubles, the training process becomes more difficult[6].

The Residual Neural Network has far fewer filters and lower complexity during training compared to VGG. The network is changed into its equivalent residual version by the addition of a shortcut connection. This shortcut connection accomplishes identity mapping with additional zero entries padding the dimensions. This choice introduces no new parameter. F(xW+x), which is the projection shortcut's mathematical representation, is utilized to match dimensions determined by 11 convolutions[6]. Fig 15. shows the architecture for ResNet 50.

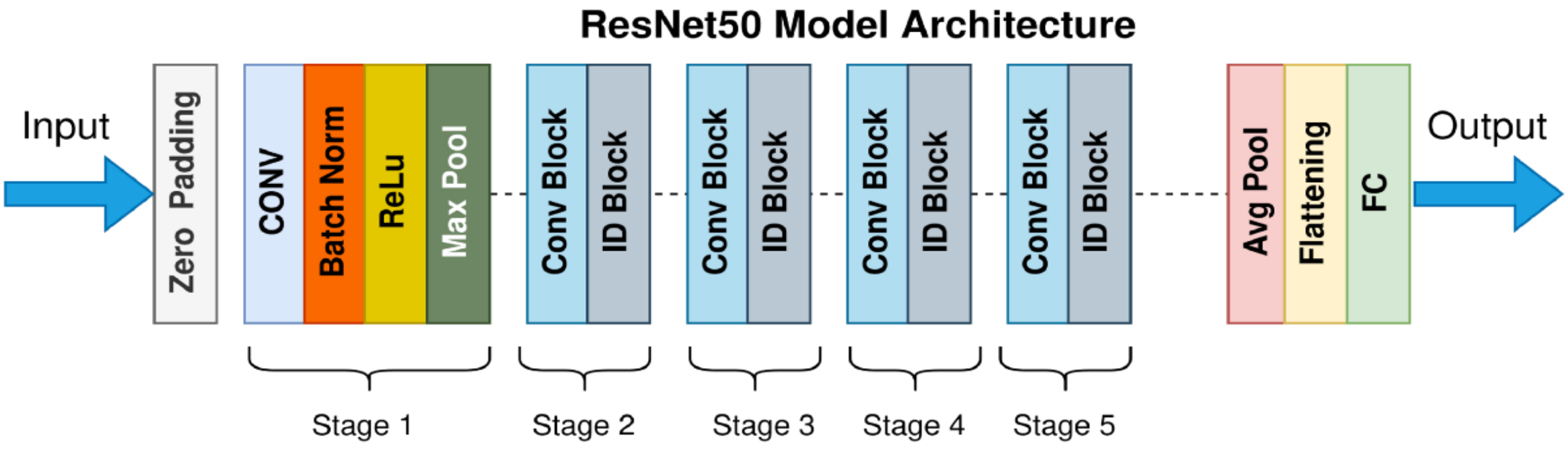

Fig 15. Resnet-50 Architecture

The pre-trained ResNet50 model is used with categorical cross entropy loss function and accuracy as metrics for the training purpose. Adam optimizer with 0.001 as the learning rate is used. The model's training and testing accuracy regarding epochs are shown below in Fig 16.

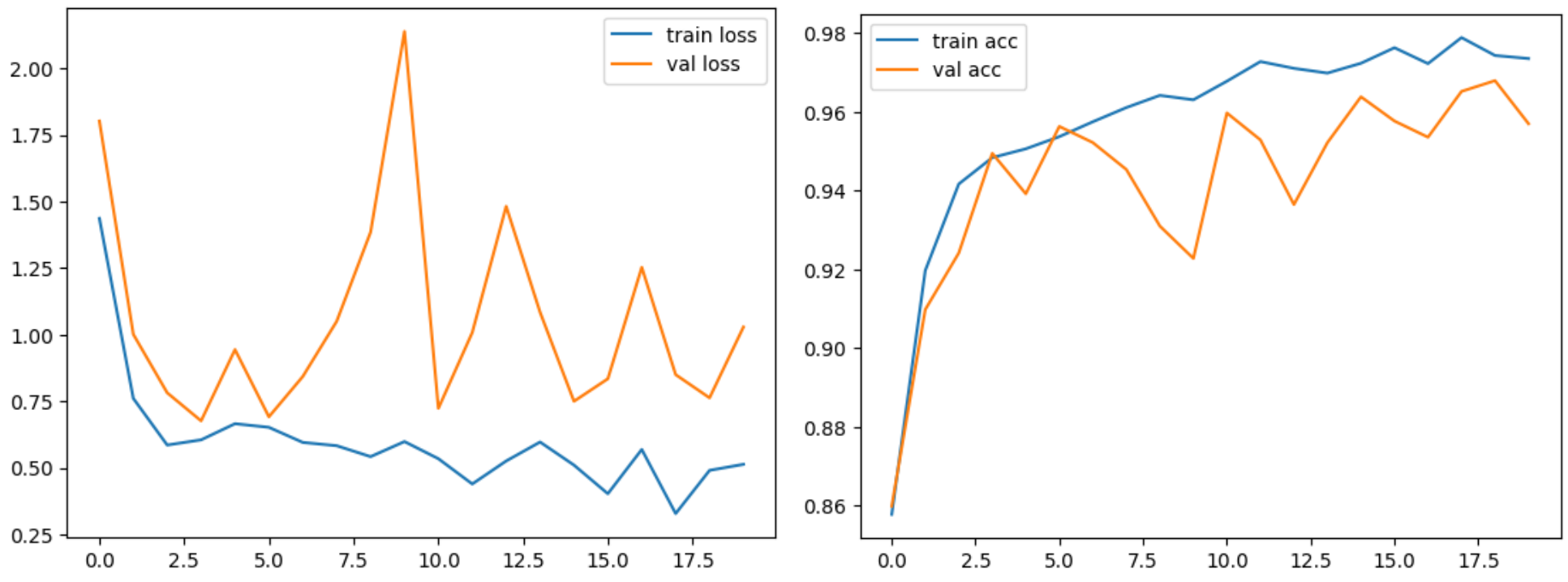

Fig 16. Training and Testing Accuracy and Loss for resNet 50

### 3.3.7 VGG 16

VGG-16 is a convolutional neural network architecture widely used in computer vision tasks, particularly image classification. It was developed by the Visual Geometry Group (VGG) at the University of Oxford in 2014. It was a runner-up in the ImageNet Large Scale Visual Recognition Challenge (ILSVRC) that same year. The network's popularity stems from its accuracy and simplicity in design. The VGG-16 architecture consists of 16 layers, including 13 convolutional layers and 3 fully connected layers. The convolutional layers use small filters with a kernel size of 3x3 and a stride of 1, while the fully connected layers have 4096 neurons each. The network is trained on the ImageNet dataset, which contains over 1 million images of 1000 classes. During training, the input images are preprocessed by subtracting the mean pixel value of the dataset from each pixel.

One of the main advantages of the VGG-16 architecture is its uniformity. All convolutional layers use the same filter size and stride, which makes it easier to interpret and understand the network. Additionally, the architecture's depth allows it to learn more complex features than shallower networks. This is particularly useful in image classification tasks, where the network needs to identify patterns at different scales.

In terms of performance, VGG-16 has achieved state-of-the-art results on several benchmark datasets, including ImageNet and CIFAR-10. Its accuracy is attributed to its depth and uniformity and its use of small filters. However, the network's main drawback is its computational cost. Training and testing VGG-16 requires a significant amount of resources, particularly memory.

Despite its computational cost, VGG-16 has been used in various computer vision tasks beyond image classification. For example, it has been applied to object detection, face recognition, and image segmentation. The network's ability to learn high-level features has made it a popular choice in transfer learning, where the network is pre-trained on a large dataset and fine-tuned on a smaller one.

In conclusion, VGG-16 is a deep convolutional neural network architecture that has significantly impacted the field of computer vision. Its uniformity and depth make it a powerful tool for image classification and other tasks, although its computational cost remains challenging. Despite this, VGG-16's popularity has led to numerous applications in a wide range of fields, from healthcare to self-driving cars[6]

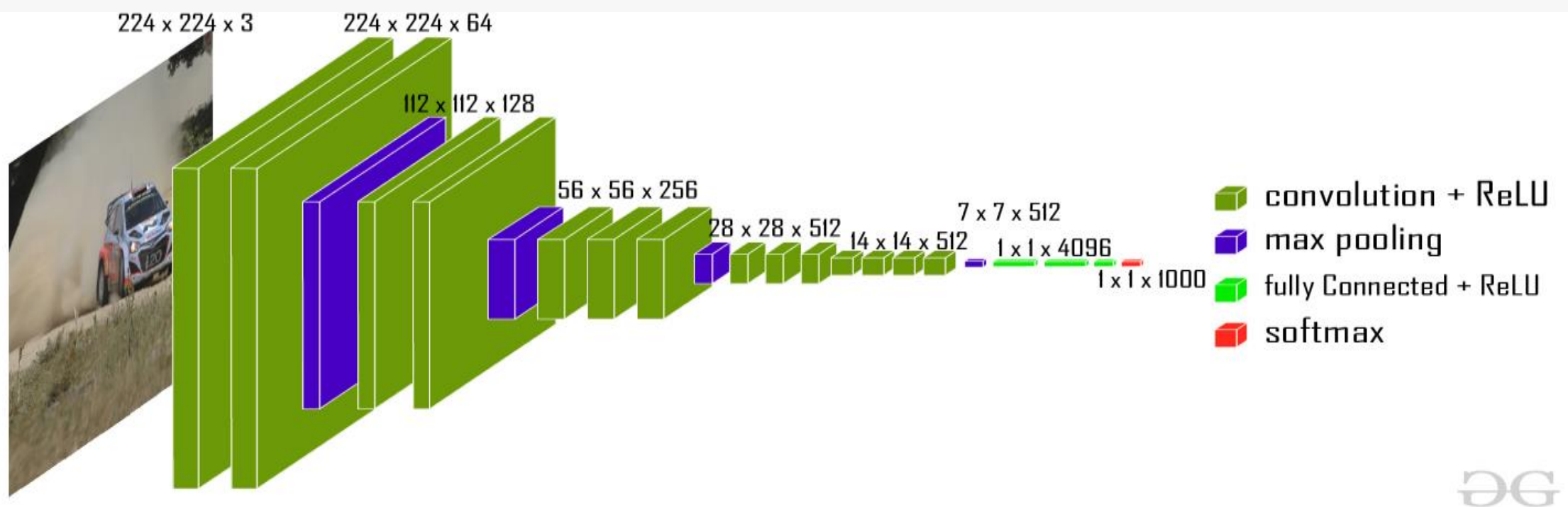

Fig 17. VGG 16 Architecture

The trained VGG16 model is utilized for training with accuracy and category cross-entropy loss function as metrics. It uses the Adam optimizer with a learning rate of 0.001. The following graphs display the training and testing accuracy of the model in Fig 18.

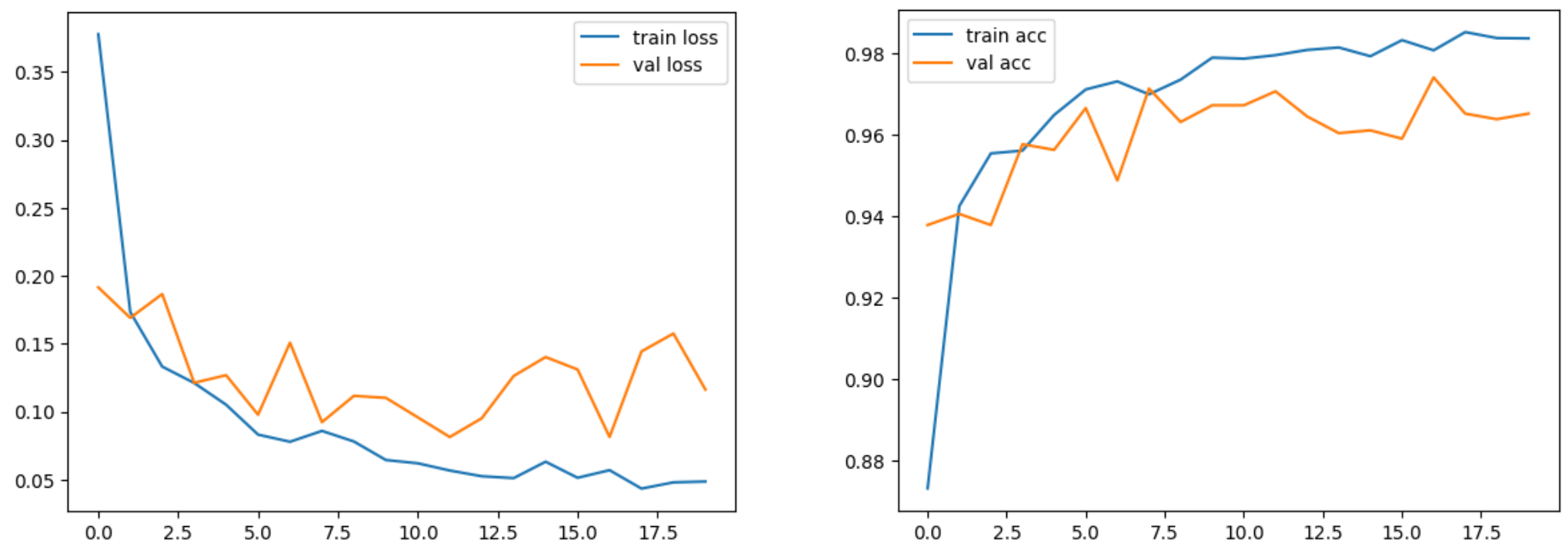

Fig 18. Training and Testing Accuracy and Loss for VGG 16

## 4. Results

Several algorithms have been applied to train models like CNN and transfer learning models like Alexnet, InceptionV3, Resnet50, and VGG16. Along with this, GAN is also implemented with CNN to train the model. The outcomes of the techniques implemented are displayed in table 1.

The detailed results of the comparison of all the models in the form of a chart are shown in Fig 19.

Fig 20. shows the predicted results of the Indian Sign language.

Table 1. Comparison of various deep learning methods w.r.t Indian Sign Language

| Model | Training Accuracy | Training Loss | Validation Accuracy | Validation Loss |
|---|---|---|---|---|
| **CNN** | 0.9328 | 0.02102 | 0.9874 | 0.0653 |
| **Alexnet** | 0.9891 | 0.0348 | 0.9969 | 0.0071 |
| **Inception V3** | 0.9735 | 0.5133 | 0.9669 | 1.028 |
| **Resnet50** | 0.9825 | 0.0540 | 0.9927 | 0.0470 |
| **VGG16** | 0.9836 | 0.0486 | 0.9851 | 0.1163 |

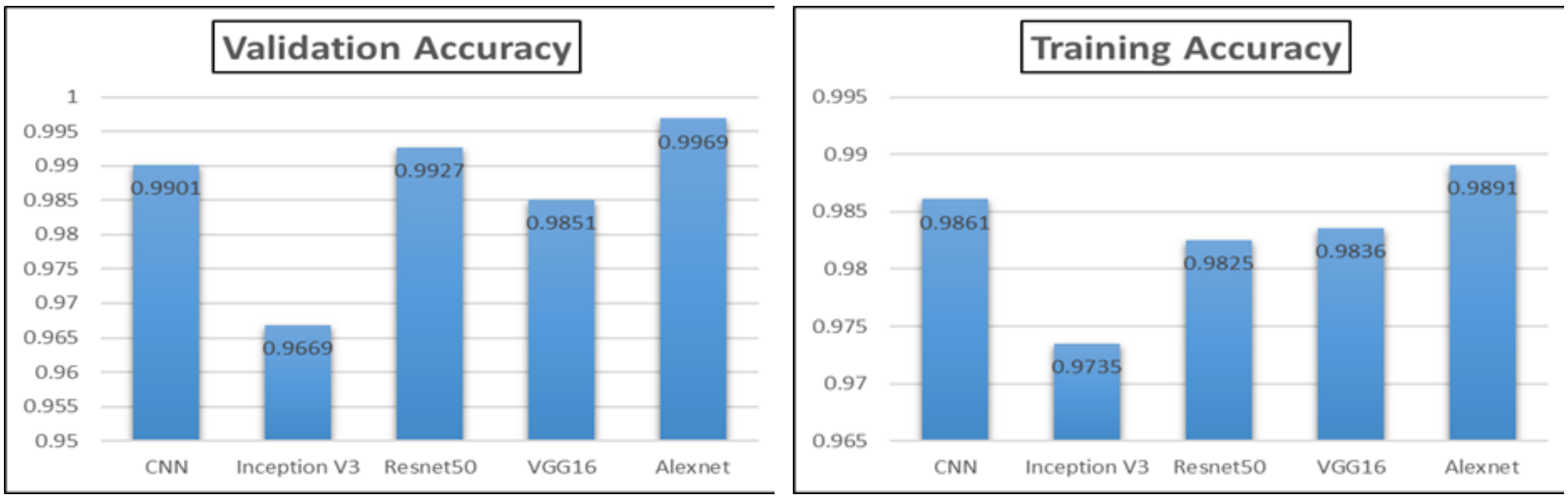

Fig 19. Accuracy comparison of various methods

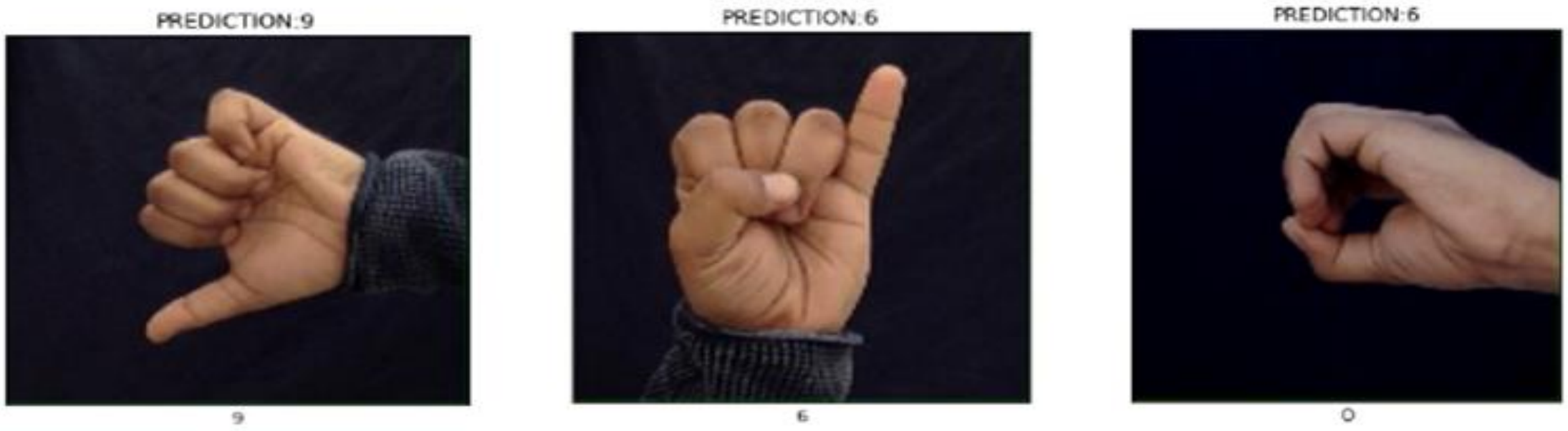

Fig 20. Prediction of Sign Language

## 5. CONCLUSION

The Sign Language Recognition (SLR) system is a way to recognize a group of created signs and translate them into text with the relevant context. Gesture recognition has essential ramifications for designing effective human-machine interactions and interactions between people with and without hearing and speech disabilities. In this study, we attempted to develop a convolutional neural network-based model. We discovered that AlexNet beat the other three pre-trained models and the traditional CNN model in terms of accuracy, resulting in 98.91% training accuracy and 99.6% validation accuracy. As a result, this method can be used to create real-time applications for sign language recognition systems in the future.

**Future Enhancements**

In the future, we will make our dataset execute word-level sign language recognition and perform its real-time implementation via results obtained through this research. In the future, this can also be employed using the web or an app to detect Indian sign language and make it simpler to be used by the deaf and hard-of-hearing people in India. Some feature extraction techniques like SURF, HOG, and SIRF can also be used to achieve higher accuracy.